%% file: Neurips_workshop/neurips_2026.tex
\documentclass{article}

\usepackage[final]{neurips_2026}
\makeatletter
\providecommand{\@trackname}{}
\makeatother
\workshoptitle{Can We Trust the Judge? Building Reliable Evaluation for Language Models}

\usepackage[utf8]{inputenc}
\usepackage[T1]{fontenc}
\usepackage{url}
\usepackage{booktabs}
\usepackage{amsfonts}
\usepackage{nicefrac}
\usepackage{microtype}
\usepackage{xcolor}
\usepackage{graphicx}
\usepackage{tabularx}
\usepackage{multirow}
\usepackage{placeins}
\usepackage{hyperref}
\usepackage{amsmath}
\usepackage{float}
\hypersetup{
  hidelinks,
  pdftitle={Does Reasoning Improve Psychological Depth in Large Language Models?},
  pdfauthor={Anonymous Authors},
  pdfsubject={Reliability of LLM-based psychological-depth evaluation}
}

\title{Does Reasoning Improve Psychological Depth in \\ Large Language Models? Depends on Who's Judging}

\author{%
  \textbf{Ruichen Zheng}\textnormal{\textsuperscript{1}} \quad
  \textbf{Yihe Wang}\textnormal{\textsuperscript{2}} \quad
  \textbf{Fabrice Y. Harel-Canada}\textnormal{\textsuperscript{3}} \quad
  \textbf{Sara Khosravi}\textnormal{\textsuperscript{1}} \\
  \textbf{Zeynep Senahan Yildiz}\textnormal{\textsuperscript{1}} \quad
  \textbf{Amit Sahai}\textnormal{\textsuperscript{1}} \quad
  \textbf{Nanyun Peng}\textnormal{\textsuperscript{4}} \\
  \vspace{0.2em} \\
  \textsuperscript{1}University of California, Los Angeles \quad
  \textsuperscript{2}Tsinghua University \\
  \textsuperscript{3}LA General Hospital \quad
  \textsuperscript{4}Google \\
}

\begin{document}

\maketitle

\input{content_workshop_edit/00_abstract}
\input{content_workshop_edit/01_introduction}
\input{content_workshop_edit/02_study_design}
\input{content_workshop_edit/03_results}
\input{content_workshop_edit/04_discussion}

{
\small
\bibliographystyle{plainnat}
\bibliography{COLM/colm2026_conference}
}

\input{content_workshop/appendix}

\end{document}

%% file: content_workshop_edit/00_abstract.tex
\begin{abstract}
LLM-as-a-Judge evaluators are increasingly used to score open-ended generation, yet a judge's correlation with human ratings on its development set may not guarantee valid measurement when outputs are closely matched and human preferences are subjective. We study this failure mode through psychological depth in short stories. Seven human readers and an LLM-judge ensemble selected on the original scalar Psychological Depth Scale dataset ($\rho = 0.646$) evaluated 60 blinded, prompt-matched story pairs from GPT-5 vs.\ GPT-4o and DeepSeek-R1 vs.\ DeepSeek-V3. Human preferences showed no universal reasoning advantage: GPT-5 was modestly preferred over GPT-4o (60.0--62.9\%), whereas DeepSeek-R1 trailed V3 (42.9\%), and inter-reader agreement was near chance (Krippendorff's $\alpha = 0.070$), with within-reader consistency and recurring weighting patterns suggesting structured heterogeneity rather than random responding. The judge, by contrast, favored reasoning outputs in 89.0\% of dimension-level comparisons and 59 of 60 pairs on aggregate PDS, uniformly across all five evaluator configurations, and its scores were associated with surface features such as sentence length and lexical diversity. These results suggest that development-set performance is insufficient evidence for deployment validity on a shifted distribution, and that point-estimate judges can obscure the heterogeneity in subjective human evaluation.
\end{abstract}

%% file: content_workshop_edit/01_introduction.tex
\section{Introduction}
\label{sec:intro}
Automatic judges are used to select checkpoints, compare systems, and increasingly provide training signals; LLM-as-a-Judge methods make this scalable, but their validity is contextual rather than intrinsic. Pairwise judging can align with human preference yet amplify evaluator biases \citep{zheng2023judgingllmasajudgemtbenchchatbot,liu2024aligning,jeong-etal-2025-comparative}, and prior work documents position, verbosity, self-preference, and inconsistency effects \citep{li-etal-2025-generation,10.5555/3737916.3740113,stureborg2024large,wang2024large}. Less is known about a more basic question: whether a judge that agrees well with humans on the data it was selected on remains valid when deployed on a shifted distribution where outputs are closely matched and human preferences are subjective.
 
Subjective creative evaluation is where this question is most acute. Fluency-centered metrics reward well-formed text but miss the qualities that matter in fiction \citep{gomez-rodriguez-williams-2023-confederacy,lu2025rethinking}, and what looks like creativity in LLM output can be surface imitation of literary style \citep{chakrabarty2024art}, so a judge may score polish rather than depth. At the same time, disagreement among readers on subjective tasks often reflects task ambiguity and differing values rather than annotation noise \citep{clark-etal-2021-thats,sandri-etal-2023-dont,kirk2024prism}. A judge evaluating such text must therefore be assessed not only on aggregate agreement but on whether it preserves the \emph{distribution} of human judgments.
 
We study this through \emph{psychological depth} in short stories, using the Psychological Depth Scale (PDS) \citep{harelcanada2024measuringpsychologicaldepthlanguage}, which measures authenticity, empathy, engagement, emotion provocation, and narrative complexity. As a test case we use reasoning-enhanced models such as GPT-5 and DeepSeek-R1, which are routinely evaluated on mathematical, coding, and factual benchmarks \citep{openai-o1,snell2024scaling} but whose reasoning-oriented post-training has rarely been examined on open-ended creative generation. %% RESTORED from COLM intro para 1; add the other benchmark cites (chen2025, li2025b, zeng2025, yu2025, plaat2025) if they are still in the .bib
We constructed 60 blinded, prompt-matched story pairs across three frontier matchups, GPT-5 versus GPT-4o under two reasoning-effort settings and DeepSeek-R1 versus DeepSeek-V3, and evaluated each pair with seven human readers and an LLM-as-a-Judge ensemble selected on the original scalar PDS dataset ($\rho = 0.646$). We ask whether a PDS judge selected for its agreement with human ratings reproduces human readers' preferences on frontier story pairs, how those preferences vary across model families, dimensions, and readers, and which story properties are associated with the judge's scores.
 
Our results expose a sharp mismatch, and we contribute three things: (1) an audit showing that a PDS judge with a respectable development-set correlation produces a near-uniform pro-reasoning signal (89.0\% of dimension-level comparisons, 59 of 60 pairs, across all five configurations) where human readers are at chance and split by model family; (2) evidence that the judge's scores track surface features such as sentence length and lexical uniformity, offering a candidate mechanism for the skew; and (3) a distributional analysis showing that reader disagreement is structured, so that a point-estimate judge cannot represent the target it was selected to approximate.

%% file: content_workshop_edit/02_study_design.tex
\section{Study design}
\label{sec:study}
 
\paragraph{Automatic evaluator setup.}
We used a scalar-rating evaluator rather than a directly pairwise-trained judge because the original PDS dataset provides scalar supervision with strong inter-annotator agreement, which let us construct calibrated few-shot prompts and compare configurations against human ratings before deployment. %% RESTORED from COLM 5.1 (replaces "validated against ... before deployment", which conflicted with the development-set framing below)
Following \citet{harelcanada2024measuringpsychologicaldepthlanguage}, the pipeline uses DSPy to enforce structured output control \citep{dspy} and retains the Mixture-of-Personas prompting strategy \citep{salewski2023context}. Using the original 97-story dataset and its human annotations, we partitioned the data into a 20\% demonstration pool and an 80\% selection split, selected representative few-shot examples from the demonstration pool by applying $k$-means clustering to the PDS score vectors \citep{su2022selective}, and used the selection split to compare Llama 3.1 70B, Llama 3.3 70B, and Qwen3.5-397B configurations across demonstration counts and Chain-of-Thought prompting \citep{grattafiori2024llama3herdmodels,qwen35blog}. No single configuration performed best across all dimensions, so we constructed a heterogeneous ensemble that assigns each PDS dimension to its best-performing configuration. The ensemble achieves $\rho=0.646$, outperforming every single-model configuration as well as the zero-shot PDS evaluator of \citet{harelcanada2024measuringpsychologicaldepthlanguage} ($\rho\approx0.51$ with GPT-4o). %% RESTORED from COLM 5.3; drop the zero-shot clause first if the page overflows
Because the same 80\% split both selected configurations and supplied this correlation, $\rho=0.646$ is a development-set estimate, mirroring how judges are commonly tuned and reused in practice. At inference, each story was scored independently on a 1--5 scale; dimension-level preference followed from the higher routed score, aggregate-PDS preference from the higher mean of the five routed scores, and exact ties contributed 0.5 to each story (Appendix~\ref{app:evaluator}).
 
\paragraph{Deployment set: frontier story pairs.}
The original PDS evaluator was calibrated on human stories and 2023-era LLM generations with quality gaps large enough to yield strong annotator consensus ($\alpha\approx0.72$); we deployed it on a considerably harder distribution. We curated 20 premises that provide contextual framing for character development while leaving the narrative direction open: 15 from r/WritingPrompts, restricted to posts after the models' training-data cutoffs, ranked by all-time upvotes, and manually filtered for quality, and 5 from Reedsy. %% RESTORED detail from COLM 3 (Premise Selection)
For each premise we generated three prompt-matched pairings: DeepSeek-R1 versus DeepSeek-V3, GPT-5 (auto effort) versus GPT-4o, and GPT-5 (high effort) versus GPT-4o, yielding 60 pairs of 400--600-word stories. All stories were generated with a common zero-shot story-writing template, with only limited family-specific wording adjustments for word-count compliance (Appendix~\ref{app:generation-details}). %% RESTORED from COLM 3; "shared system prompt" glossed over the DeepSeek length sentence
DeepSeek-R1 is an RL upgrade of V3 and provides our closest anchor for reasoning-oriented post-training \citep{Guo_2025,liu2024deepseek}; the GPT pairs are model-level contrasts \citep{singh2025openaigpt5card,hurst2024gpt}. These are ecologically relevant comparisons rather than clean ablations of reasoning, and we use ``reasoning output'' as a label for the model, not a causal attribution.
 
\paragraph{Human evaluation.}
We adapted the \citet{harelcanada2024measuringpsychologicaldepthlanguage} framework to a blinded pairwise protocol; the Institutional Review Board deemed the study exempt. %% RESTORED from COLM 4
Seven undergraduate readers from English and Psychology departments were onboarded on the five PDS dimensions and a custom Label Studio annotation interface \citep{LabelStudio}. For each of the 60 pairs, the two stories were presented side by side in randomized order, with the PDS rubric definitions shown at the top of the page, and readers provided six forced-choice judgments: one for each of the five PDS dimensions and one overall preference, with no tie or confidence option. This produced $7\times60\times6=2{,}520$ pairwise preference judgments along with 112 reader-produced free-text justifications. To monitor annotation quality, we logged per-pair completion times, conducted a leave-one-reader-out stability analysis, and administered a post-study survey; recruitment, compensation, and diagnostics are in Appendix~\ref{app:human-diagnostics}. %% RESTORED from COLM 4.2 with "annotators" -> "readers"
 
\paragraph{Analysis.}
Human preference intervals come from separate intercept-only logistic mixed models per matchup $\times$ criterion, $\mathrm{logit}\,P(y_{ir}=1)=\beta_0+u_i+v_r$, with crossed random intercepts for premise pair $i$ and reader $r$; the 18 intervals are pointwise and unadjusted for multiplicity. We report Fleiss' $\kappa$ and Krippendorff's $\alpha$ \citep{fleiss1971measuring, krippendorff2011computing} as agreement summaries, and per-reader logistic regressions and dimension-to-overall Cohen's $\kappa$ to characterize how readers weight dimensions; with seven readers these are exploratory. To probe surface associations, we measured length, sentence-length moments, MATTR, MTLD, and perplexity under Llama 3.1 70B. Full results appear in Appendices~\ref{app:glmm} and~\ref{app:surface}.

%% file: content_workshop_edit/03_results.tex
\section{Results}
\label{sec:results}
 
\begin{figure}[t]
  \centering
  \begin{minipage}[b]{0.555\linewidth}
    \centering
    \includegraphics[width=\linewidth]{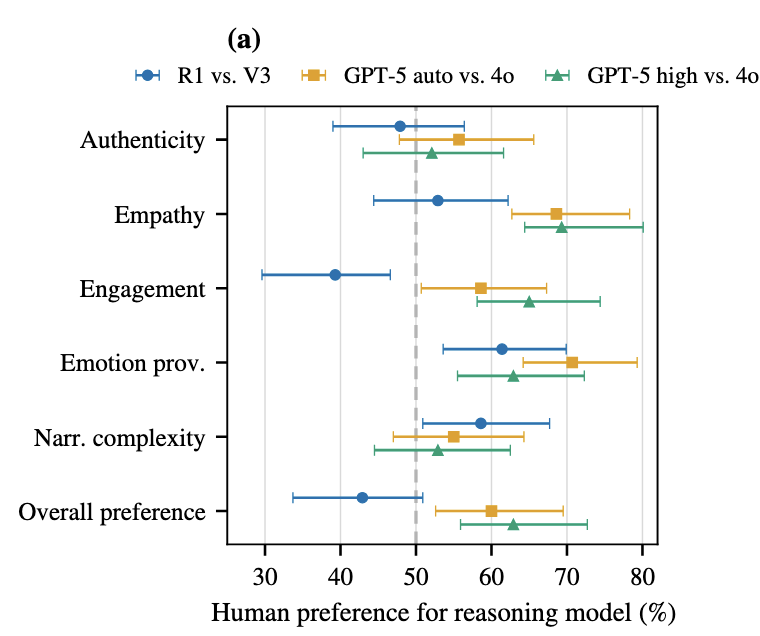}
  \end{minipage}\hfill
  \begin{minipage}[b]{0.435\linewidth}
    \centering
    \includegraphics[width=\linewidth]{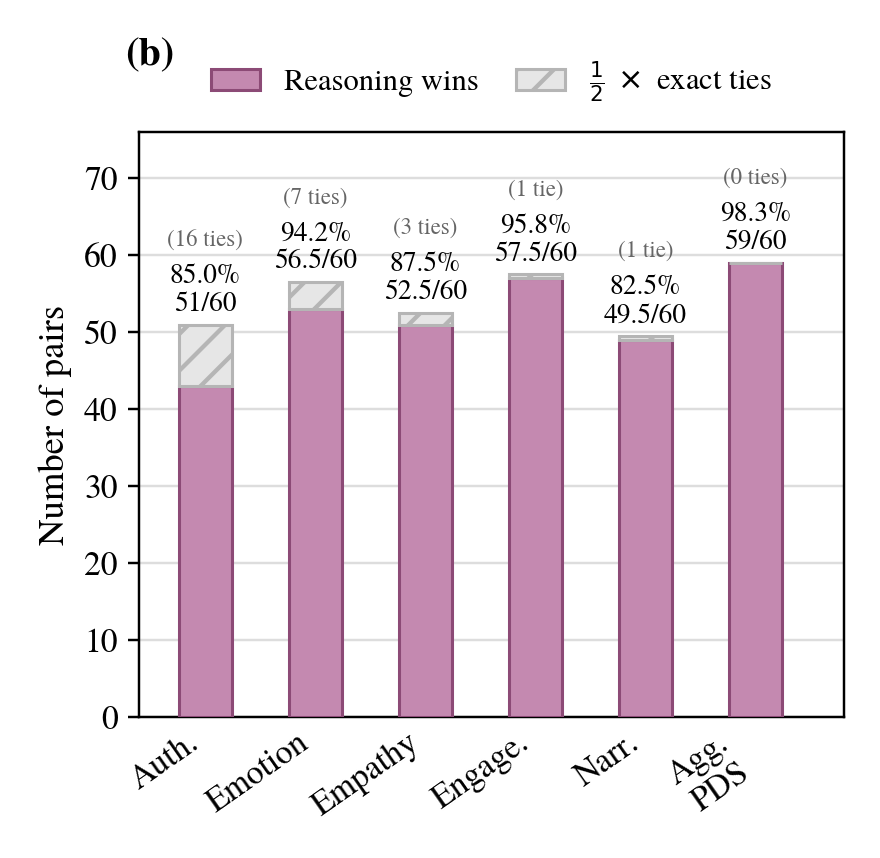}
  \end{minipage}
  \caption{\textbf{Human preferences vary by model pairing and dimension; the LLM judge does not.} \textbf{(a)} Human preference for the reasoning-model story with pointwise 95\% GLMM intervals; the dashed line marks chance. \textbf{(b)} LLM-judge preference for the reasoning-model story pooled over all 60 pairs. Solid bars are reasoning wins; hatched segments credit each exact score tie 0.5 to the reasoning story; labels give the resulting even-split rate, with raw tie counts in parentheses. Aggregate PDS averages the five dimension scores and is a distinct target from the human overall-preference question.}
  \label{fig:compact-results}
\end{figure}
 
\paragraph{The judge is nearly uniformly pro-reasoning; readers are not.}
Across the five PDS dimensions, the ensemble favored the reasoning-model story in 89.0\% of 300 pair-by-dimension comparisons with ties split evenly (Figure~\ref{fig:compact-results}b), ranging from 82.5\% on narrative complexity to 95.8\% on engagement. On aggregate PDS it selected the reasoning story in 59 of 60 pairs: all 40 GPT-family pairs and, despite readers leaning toward V3, 19 of 20 DeepSeek pairs. Human readers, by contrast, gave reasoning stories 58.0\% of 2,100 dimension-level votes and 55.2\% of 420 overall-preference votes. This skew is not an artifact of the ensemble construction: all five individual judge configurations preferred the reasoning model, with pooled even-split rates from 77.5\% to 96.3\% (Appendix~\ref{app:evaluator}). %% RESTORED phrasing from COLM 6.3
 
\paragraph{Human preference depends on model family and dimension.}
Human preferences diverge across model families (Figure~\ref{fig:compact-results}a). Although reasoning stories received 55.2\% of overall-preference votes, this aggregate obscures a family-level split: readers modestly preferred GPT-5 over GPT-4o under both the auto (60.0\%, 95\% CI [52.6, 69.5]) and high (62.9\%, [55.9, 72.7]) reasoning-effort settings, whereas DeepSeek-R1 trailed DeepSeek-V3 (42.9\%, [33.7, 50.9]), an interval that includes chance. The same divergence appears across PDS dimensions: the GPT-5 pairings were consistently above 50\% on the affective dimensions, especially emotion provocation (70.7\%) and empathy (68.6\%), whereas R1 vs.\ V3 exhibited mixed effects, notably disfavored on engagement (39.3\%, [29.6, 46.6]) yet favored on emotion provocation (61.4\%) and narrative complexity (58.6\%). Individual readers ranged from strongly pro-reasoning (Reader~7, 80.0\%) to pro-base (Reader~1, 36.7\%), so the pooled 55.2\% masks both family- and reader-level variation (Appendix~\ref{app:glmm}). %% RESTORED structure and phrasing from COLM 6.1 with GLMM intervals inserted
 
\paragraph{Group agreement is low; reader heterogeneity is structured but exploratory.}
Inter-reader agreement is low: Krippendorff's $\alpha=0.070$, with raw pairwise agreement (53.8\%) barely above the 50\% expected by chance (Appendix~\ref{app:glmm}). This is a dramatic drop from the $\alpha\approx0.72$ reported by \citet{harelcanada2024measuringpsychologicaldepthlanguage} using the same rubric on large-gap items. Our frontier pairs are far closer in quality, and the low agreement may reflect limited rubric discrimination on close pairs, reader subjectivity, or noise. Quality checks reduce but do not eliminate annotation-failure concerns: readers spent a median of 108--418 seconds per pair with stable pace throughout, and removing any single reader changes the aggregate rate by only 2--4 points (Appendix~\ref{app:human-diagnostics}). Per-reader regressions, dimension-to-overall $\kappa$, and free-text rationales show recurring dimension-weighting patterns, consistent with structured heterogeneity rather than random responding, though seven readers preclude definitive clusters (Appendix~\ref{app:glmm}).
 
\paragraph{Surface features are associated with judge scores.}
Judge engagement scores correlated with mean sentence length at $r=-0.70$, and moving-average type-token ratio was negatively associated with judge scores on four of five dimensions ($-0.37$ to $-0.14$); reasoning-model stories had shorter sentences and slightly lower MATTR (0.896 vs.\ 0.907), so the judge's preference is partly associated with these properties rather than depth alone. Reasoning stories were also more perplexing to Llama~3.1~70B, arguing against a simple familiarity account (Appendix~\ref{app:surface}).

%% file: content_workshop_edit/04_discussion.tex
\section{Implications and limitations}
\label{sec:implications}
 
\paragraph{Development-set correlation is not deployment validity.}
The ensemble agreed well with human PDS ratings on its selection set ($\rho=0.646$), yet on frontier pairs it produced a near-uniform pro-reasoning signal where readers were at chance. Because selection and validation shared the same split, this is a context-specific audit rather than independent evidence of transfer failure. All five configurations agreed with each other far more than seven readers agreed among themselves, so inter-judge consistency \citep{li-etal-2025-generation,stureborg2024large} offered no warning. This is all the more notable because most of our evaluator models predate the reasoning era and do not plausibly favor reasoning outputs through training-data exposure to synthetic reasoning traces; the association of judge scores with sentence length and lexical uniformity points instead to sensitivity to stylistic artifacts of reasoning-oriented generation. Evaluator reports should document selection procedure, deployment context, and known surface sensitivities, and include a target-matched human check on the deployment distribution.
 
\paragraph{Limitations.}
This exploratory audit uses 20 premises, one retained generation per condition, seven readers from one university, and one domain. Forced choice without ties or confidence may turn indifference into preference; limited PDS discrimination on close pairs, fatigue, and noise may all contribute to low agreement. The human target changed and surface results are correlational. Broader cohorts, repeated generations, target-matched judges, and out-of-fold validation are needed.
 
\section{Conclusion}
The PDS judge that agreed well with human ratings on its selection set was nearly uniformly pro-reasoning on closely matched frontier story pairs where human readers were at chance and split by model family. Our exploratory audit motivates target-matched human checks and independent validation before reusing a selected evaluator on a new distribution.

%% file: content_workshop/appendix.tex
\clearpage
\appendix

\section{Full human-preference results}
\label{app:glmm}
\label{app:reader-heterogeneity}

This section expands the human-preference and reader-heterogeneity findings in Section~\ref{sec:results}. We fitted a separate intercept-only logistic GLMM to each matchup $\times$ criterion, with reasoning-model choice coded as 1 and crossed random intercepts for premise pair and reader:
\[
\mathrm{logit}\,P(y_{ir}=1)=\beta_0+u_i+v_r.
\]
Table~\ref{tab:glmm-preferences} reports the resulting preference estimates and 95\% intervals summarized in Figure~\ref{fig:compact-results}. Table~\ref{tab:agreement} reports inter-reader agreement by dimension, and Table~\ref{tab:agreement-by-pairing} breaks Krippendorff's $\alpha$ down by model pairing; agreement is weak in every matchup, with several near-zero or negative cells, so the low group agreement is not driven by any single comparison.

\begin{table}[!htbp]
\caption{Human preference for the reasoning-model output, with pointwise 95\% GLMM intervals. Rates above 50\% favor the reasoning model. The 18 intervals are unadjusted for multiplicity and interpreted descriptively.}
\label{tab:glmm-preferences}
\centering
\small
\begin{tabular}{lccc}
\toprule
& \textbf{DeepSeek-R1} & \textbf{GPT-5 auto} & \textbf{GPT-5 high} \\
\textbf{Dimension} & \textbf{vs.\ V3} & \textbf{vs.\ 4o} & \textbf{vs.\ 4o} \\
\midrule
Authenticity
& \shortstack{47.9\%\\{[39.0, 56.4]}}
& \shortstack{55.7\%\\{[47.8, 65.6]}}
& \shortstack{52.1\%\\{[43.0, 61.6]}} \\
Empathy
& \shortstack{52.9\%\\{[44.4, 62.2]}}
& \shortstack{68.6\%\\{[62.7, 78.3]}}
& \shortstack{69.3\%\\{[64.4, 80.1]}} \\
Engagement
& \shortstack{39.3\%\\{[29.6, 46.6]}}
& \shortstack{58.6\%\\{[50.7, 67.3]}}
& \shortstack{65.0\%\\{[58.1, 74.4]}} \\
Emotion prov.
& \shortstack{61.4\%\\{[53.6, 69.9]}}
& \shortstack{70.7\%\\{[64.2, 79.3]}}
& \shortstack{62.9\%\\{[55.5, 72.3]}} \\
Narr.\ complexity
& \shortstack{58.6\%\\{[50.9, 67.7]}}
& \shortstack{55.0\%\\{[47.0, 64.3]}}
& \shortstack{52.9\%\\{[44.5, 62.5]}} \\
\midrule
\textbf{Overall preference}
& \shortstack{42.9\%\\{[33.7, 50.9]}}
& \shortstack{60.0\%\\{[52.6, 69.5]}}
& \shortstack{62.9\%\\{[55.9, 72.7]}} \\
\bottomrule
\end{tabular}
\end{table}

\begin{table}[!htbp]
\caption{Inter-reader agreement across 60 pairs and seven readers. Raw is mean pairwise agreement.}
\label{tab:agreement}
\centering
\small
\begin{tabular}{lccc}
\toprule
\textbf{Dimension} & \textbf{Fleiss' $\kappa$} & \textbf{Kripp.\ $\alpha$} & \textbf{Raw} \\
\midrule
Authenticity & 0.035 & 0.037 & 52.5\% \\
Emotion prov. & 0.080 & 0.082 & 54.1\% \\
Empathy & 0.182 & 0.185 & 59.1\% \\
Engagement & 0.059 & 0.061 & 53.3\% \\
Narr.\ complexity & 0.016 & 0.018 & 52.1\% \\
Overall preference & 0.035 & 0.038 & 51.9\% \\
\midrule
\textbf{Average} & 0.068 & 0.070 & 53.8\% \\
\bottomrule
\end{tabular}
\end{table}

\begin{table}[!htbp]
\caption{Krippendorff's $\alpha$ by model pairing.}
\label{tab:agreement-by-pairing}
\centering
\small
\begin{tabular}{lcccc}
\toprule
\textbf{Dimension} & \textbf{All} & \textbf{R1 vs.\ V3} & \textbf{GPT-5 auto vs.\ 4o} & \textbf{GPT-5 high vs.\ 4o} \\
\midrule
Authenticity & 0.037 & 0.073 & 0.053 & $-0.015$ \\
Emotion prov. & 0.082 & 0.035 & 0.147 & 0.072 \\
Empathy & 0.185 & 0.047 & 0.176 & 0.290 \\
Engagement & 0.061 & 0.101 & 0.010 & 0.080 \\
Narr.\ complexity & 0.018 & 0.049 & $-0.013$ & 0.005 \\
Overall preference & 0.038 & 0.074 & $-0.003$ & 0.050 \\
\midrule
\textbf{Average} & 0.070 & 0.063 & 0.062 & 0.080 \\
\bottomrule
\end{tabular}
\end{table}

As a dependence sensitivity check, cluster bootstrap analyses produced 10 significant model-by-dimension cells when resampling pairs and five when resampling readers, compared with 11 under the crossed-random-intercept GLMM. The smaller reader-cluster count underscores the uncertainty induced by a seven-reader sample; we therefore emphasize effect patterns and intervals rather than a binary significance tally.

\begin{figure}[!htbp]
\centering
\includegraphics[width=\linewidth]{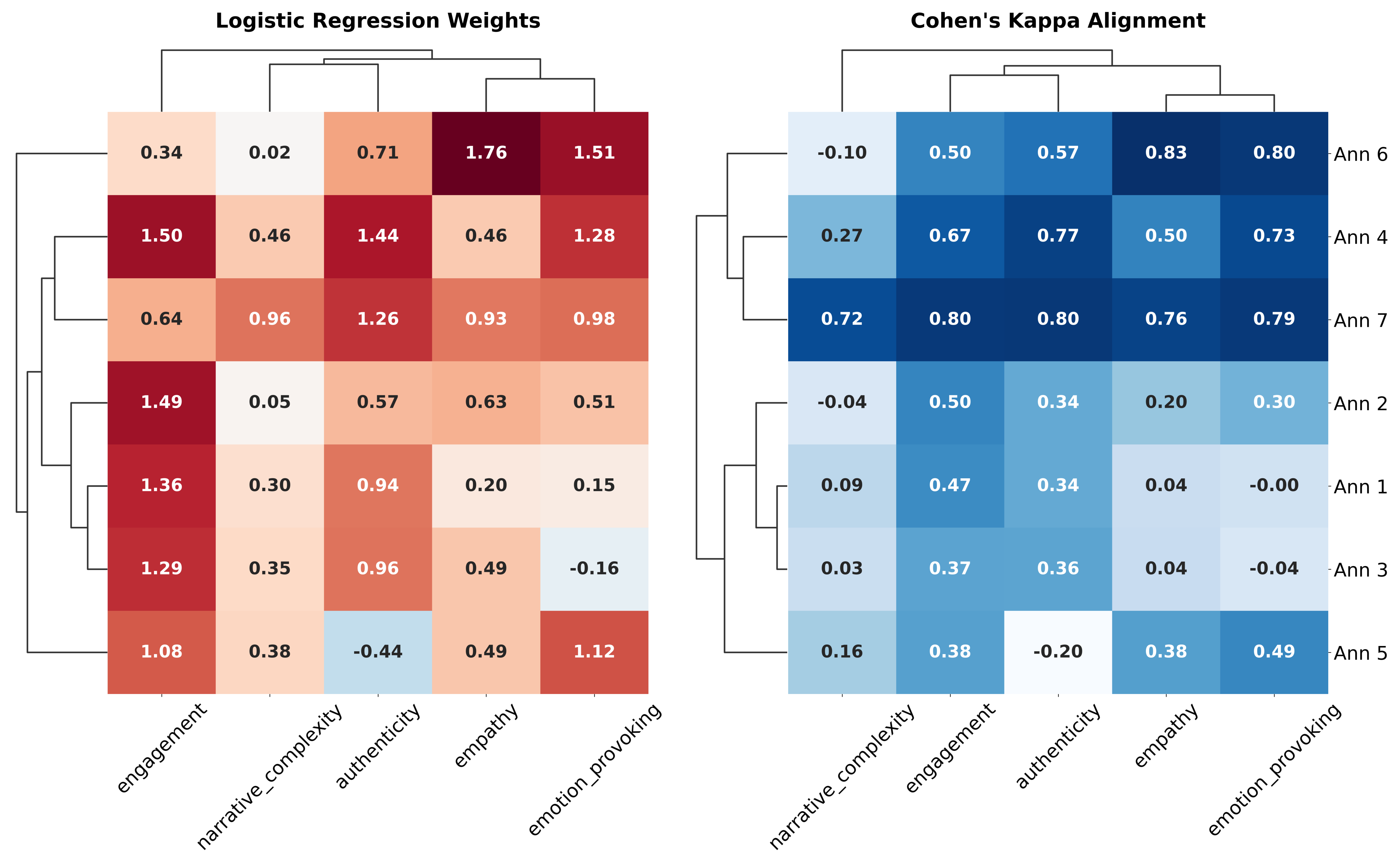}
\caption{Exploratory reader-level structure. \textbf{Left:} coefficients from separate logistic regressions relating PDS dimensions to each reader's overall preference. \textbf{Right:} within-reader Cohen's $\kappa$ between each dimension and overall preference. Similar clusters across analyses are consistent with recurring dimension-weighting patterns, but the seven-reader sample precludes population-level archetype claims.}
\label{fig:reader-clustering}
\end{figure}

Figure~\ref{fig:reader-clustering} shows the exploratory reader-level structure. Hierarchical clustering of per-reader logistic-regression coefficients and of dimension-to-overall Cohen's $\kappa$ yield similar structures with the same innermost sub-clusters, ((Reader~1, Reader~3), Reader~2) and (Reader~4, Reader~7). Readers~1 and~3 rely comparatively more on engagement and authenticity, whereas Readers~4 and~7 assign broadly positive weights across dimensions; empathy and emotion provocation are the most closely clustered dimensions. Higher within-reader $\kappa$ generally corresponds to larger regression weight. Free-text rationales (Table~\ref{tab:selected-comments}) show readers prioritizing different qualities, such as narrative clarity over emotional immediacy, rather than overlooking the same features. These patterns are consistent with recurring dimension-weighting schemes, but seven readers preclude population-level claims.

\FloatBarrier
\section{Evaluator construction and robustness}
\label{app:evaluator}

This section expands the automatic-evaluator method in Section~\ref{sec:study} and the human and judge comparison in Section~\ref{sec:results}. We used DSPy for structured scalar outputs \citep{dspy} and retained the original PDS mixture-of-personas strategy, motivated by evidence that in-context impersonation changes model behavior \citep{salewski2023context}. We selected diverse demonstrations by applying $k$-means to PDS score vectors \citep{su2022selective}. Candidate Llama and Qwen configurations \citep{grattafiori2024llama3herdmodels,qwen35blog} varied in demonstration count and Chain-of-Thought prompting \citep{wei2022chain}.

Table~\ref{tab:evaluator-selection} gives development-set correlations for the candidate configurations. The same 80\% split was used both to compare configurations and to route each PDS dimension to its highest-correlation configuration; the routed $\rho=0.646$ is consequently a selection-set estimate, not performance on an untouched test set. Its value is computed over routed predictions rather than as the arithmetic mean of the rounded dimension entries. At deployment, judge aggregate-PDS preference compares the arithmetic mean of the five routed scores,
\[
S_{\mathrm{agg}}(x)=\frac{1}{5}\sum_{d=1}^{5}S_d(x),\qquad
\widehat y_{\mathrm{agg}}=\mathbb{1}[S_{\mathrm{agg}}(R)>S_{\mathrm{agg}}(B)].
\]
All dimensions use a 1\textendash5 scale, but scores from the different routed configurations were not otherwise calibrated before averaging. This aggregate is not equivalent to the separate overall-preference question answered by human readers. Table~\ref{tab:judge-effect-sizes} reports, by pairing, the share of human overall-preference votes and the number of aggregate-PDS pair decisions favoring the reasoning story, together with the mean judge-score difference and its standardized effect size. Ties are rare for most ensemble dimensions but frequent on authenticity (16) and substantially more frequent for some individual configurations, most notably the Chain-of-Thought configuration (75). Table~\ref{tab:ties} gives per-dimension tie counts and even-split rates for the routed ensemble; the aggregate-PDS decision involves no ties.

\begin{table}[!htbp]
\caption{Human overall-preference votes and judge aggregate-PDS pair decisions favoring reasoning-model outputs. These columns measure different targets and are not direct agreement statistics. $\Delta$ is the descriptive mean paired judge-score difference (reasoning minus comparison) pooled across the five routed dimensions, and $d_z$ is the corresponding standardized mean difference. Because dimensions from the same pair are dependent, we do not report naive paired-test inference.}
\label{tab:judge-effect-sizes}
\centering
\small
\begin{tabular}{lcccc}
\toprule
\textbf{Pairing} &
\textbf{Human overall votes} &
\textbf{Judge agg.\ PDS} &
\textbf{$\Delta$} &
\textbf{$d_z$} \\
\midrule
GPT-5 auto vs.\ 4o
& 84/140 (60.0\%)
& 20/20
& $+0.518$
& 1.14 \\
GPT-5 high vs.\ 4o
& 88/140 (62.9\%)
& 20/20
& $+0.500$
& 1.12 \\
DeepSeek-R1 vs.\ V3
& 60/140 (42.9\%)
& 19/20
& $+0.214$
& 0.68 \\
\midrule
\textbf{Overall}
& \textbf{232/420 (55.2\%)}
& \textbf{59/60}
& \textbf{$+0.411$}
& \textbf{0.95} \\
\bottomrule
\end{tabular}
\end{table}

\begin{table}[!htbp]
\caption{Spearman correlation with human PDS annotations on the 80\% development/selection split. Bold marks the configuration selected for each dimension; these are not untouched-test estimates. For individual configurations, the final column is the mean of the five displayed correlations; for the routed ensemble, it is computed from routed predictions.}
\label{tab:evaluator-selection}
\centering
\small
\begin{tabular}{lcccccc}
\toprule
\textbf{Evaluator} & \textbf{AUT} & \textbf{PROV} & \textbf{EMP} & \textbf{ENG} & \textbf{NCOM} & \textbf{Mean/routed $\rho$} \\
\midrule
Llama 3.1 70B, 7 demo & 0.576 & 0.674 & 0.637 & 0.507 & 0.637 & 0.606 \\
Llama 3.1 70B, 5 demo & \textbf{0.588} & 0.655 & 0.596 & 0.450 & 0.575 & 0.573 \\
Llama 3.1 70B, 10 demo & 0.561 & \textbf{0.702} & 0.671 & 0.387 & 0.568 & 0.578 \\
Llama 3.3 70B, 10 demo & 0.580 & 0.676 & \textbf{0.702} & 0.407 & 0.550 & 0.583 \\
Qwen 3.5 397B, 7 demo & 0.511 & 0.663 & 0.602 & \textbf{0.564} & 0.496 & 0.567 \\
Llama 3.1 70B CoT, 10 demo & 0.499 & 0.627 & 0.658 & 0.356 & \textbf{0.664} & 0.561 \\
\midrule
\textbf{Routed ensemble} & 0.588 & 0.702 & 0.702 & 0.564 & 0.664 & 0.646 \\
\bottomrule
\end{tabular}
\end{table}

Individual configurations also favored reasoning-model outputs (Table~\ref{tab:individual-judges}), showing that the headline skew is not created solely by routing. These are five configurations, not independent evaluator replications; three use Llama~3.1~70B. Stars identify routed dimensions. Each dimension cell reports reasoning wins over non-tied pairs; the Decisive column pools wins over all non-tied pair-by-dimension comparisons, and the Even-split column, which the main text uses, credits each tie 0.5 to both stories over all 300 comparisons.

\begin{table}[!htbp]
\caption{Reasoning-output preference for five evaluator configurations and the routed ensemble. Dimension cells report reasoning wins / non-tied pairs; $^*$ marks the routed dimension. Decisive pools wins over non-tied comparisons; Even-split credits ties 0.5 to each story over all 300 comparisons.}
\label{tab:individual-judges}
\centering
\small
\begin{tabularx}{\linewidth}{Xcccccccc}
\toprule
\textbf{Configuration} & \textbf{Auth.} & \textbf{Emo.} & \textbf{Emp.} & \textbf{Eng.} & \textbf{Narr.} & \textbf{Ties} & \textbf{Decisive} & \textbf{Even-split} \\
\midrule
Llama 3.1 70B (5d)  & 43/44$^*$ & 54/57 & 49/54 & 47/51 & 53/55 & 39 & 94.3\% & 88.5\% \\
Llama 3.1 70B (10d) & 43/43 & 53/53$^*$ & 48/51 & 46/46 & 54/54 & 53 & 98.8\% & 90.2\% \\
Llama 3.3 70B (10d) & 32/33 & 53/55 & 51/57$^*$ & 47/58 & 47/56 & 41 & 88.8\% & 83.5\% \\
Qwen 3.5 397B (7d)  & 57/57 & 58/58 & 55/58 & 57/59$^*$ & 57/58 & 10 & 97.9\% & 96.3\% \\
Llama 3.1 70B CoT   & 7/8 & 49/50 & 47/53 & 43/55 & 49/59$^*$ & 75 & 86.7\% & 77.5\% \\
\midrule
\textbf{Routed ensemble} & 43/44 & 53/53 & 51/57 & 57/59 & 49/59 & 28 & 93.0\% & 89.0\% \\
\bottomrule
\end{tabularx}
\end{table}

\begin{table}[!htbp]
\caption{Tie frequency and reasoning-output preference for the routed ensemble. Even-split rate $=(\text{wins}+0.5\times\text{ties})/60$ per dimension, or $/300$ for the pooled row. Aggregate PDS is a separate evaluator target, not human overall preference.}
\label{tab:ties}
\centering
\small
\begin{tabular}{lccc}
\toprule
\textbf{Dimension} & \textbf{Ties} & \textbf{Reasoning wins} & \textbf{Even-split rate} \\
\midrule
Authenticity      & 16 & 43 & 85.0\% \\
Emotion prov.     &  7 & 53 & 94.2\% \\
Empathy           &  3 & 51 & 87.5\% \\
Engagement        &  1 & 57 & 95.8\% \\
Narr.\ complexity &  1 & 49 & 82.5\% \\
\textbf{Five dimensions pooled} & \textbf{28} & \textbf{253} & \textbf{89.0\%} \\
\midrule
Aggregate PDS     &  0 & 59 & 98.3\% \\
\bottomrule
\end{tabular}
\end{table}

\FloatBarrier
\section{Surface-feature analysis}
\label{app:surface}

This section provides the exploratory surface-feature associations summarized in Sections~\ref{sec:results} and~\ref{sec:implications}. We measured deterministic surface properties for all 120 story presentations; if GPT-4o texts were reused, the corresponding rows are repeated texts rather than independent outputs. For human overall preference, the pair-level mean-sentence-length difference was associated with reasoning-model preference across all 60 pairs ($r=-0.59$), with a similar descriptive association within the 40 GPT-family pairs ($r=-0.55$). DeepSeek pairs clustered near zero sentence-length difference, with no detectable association. Table~\ref{tab:surface-story} relates four representative lexical and length features to ensemble scores; Table~\ref{tab:surface-pair} repeats the analysis after within-pair standardization to remove premise-level differences. Other sentence-count moments and perplexity were used as diagnostics rather than included in the displayed feature-by-dimension coefficients.

\begin{table}[!htbp]
\caption{Story-level Pearson correlation between surface features and ensemble judge scores ($n=120$).}
\label{tab:surface-story}
\centering
\small
\begin{tabular}{lccccc}
\toprule
\textbf{Feature} & \textbf{Auth.} & \textbf{Emo.} & \textbf{Emp.} & \textbf{Eng.} & \textbf{Narr.} \\
\midrule
Word count & $-.28$ & $-.19$ & $-.12$ & $-.18$ & $+.09$ \\
Mean sentence length & $-.32$ & $-.43$ & $-.17$ & $-.70$ & $-.09$ \\
MATTR & $-.37$ & $-.33$ & $-.35$ & $-.29$ & $-.14$ \\
MTLD & $-.24$ & $-.19$ & $-.25$ & $-.09$ & $+.03$ \\
\bottomrule
\end{tabular}
\end{table}

\begin{table}[!htbp]
\caption{Within-pair standardized regression coefficients ($\beta$) relating surface features to ensemble judge scores ($n=60$ pairs).}
\label{tab:surface-pair}
\centering
\small
\begin{tabular}{lccccc}
\toprule
\textbf{Feature} & \textbf{Auth.} & \textbf{Emo.} & \textbf{Emp.} & \textbf{Eng.} & \textbf{Narr.} \\
\midrule
Word count & $-.64$ & $-.40$ & $-.18$ & $-.54$ & $+.11$ \\
Mean sentence length & $-.56$ & $-.47$ & $-.27$ & $-.59$ & $+.03$ \\
MATTR & $-.31$ & $-.17$ & $-.24$ & $-.28$ & $+.13$ \\
MTLD & $-.39$ & $-.13$ & $-.15$ & $-.36$ & $+.31$ \\
\bottomrule
\end{tabular}
\end{table}

Tables~\ref{tab:surface-story} and~\ref{tab:surface-pair} report exploratory descriptive coefficients. We omit independent-cell inference because observations share premises, model families, and potentially repeated GPT-4o texts; the coefficients are associations, not causal effects. Reasoning-model stories were slightly more locally uniform (MATTR 0.896 versus 0.907) but had higher perplexity under Llama 3.1 70B (GPT-5: 16.3 versus GPT-4o: 9.5; R1: 8.9 versus V3: 5.1). These patterns argue against simple familiarity while showing surface sensitivity.

\input{content_workshop/appendix_generation}
\input{content_workshop/appendix_annotation}
\input{content_workshop/appendix_visualizations}

%% file: content_workshop/appendix_generation.tex
\FloatBarrier
\section{Story Generation Details}
\label{app:generation-details}
\label{app:generation}

\subsection{Generation Prompt}

All models received the same system prompt, adapted from \citet{harelcanada2024measuringpsychologicaldepthlanguage}:

\begin{quote}
\small
You are a seasoned writer who has won several accolades for your emotionally rich stories. When you write, you delve deep into the human psyche, pulling from the reservoir of universal experiences that every reader, regardless of their background, can connect to. Your writing is renowned for painting vivid emotional landscapes, making readers not just observe but truly feel the world of your characters. Every piece you produce aims to draw readers in, encouraging them to reflect on their own lives and emotions. Your stories are a complex tapestry of relationships, emotions, and conflicts, each more intricate than the last.
\end{quote}

The prompt varied slightly across model families to accommodate word-count compliance:

\paragraph{GPT-4o and GPT-5.}
\begin{quote}
\small
Please write a 500-word story \{on / following this instruction\}:\\
\{premise\}\\
Only respond with the story text.
\end{quote}

\paragraph{DeepSeek-V3 and DeepSeek-R1.}
\begin{quote}
\small
Please write a 500-word story \{on / following this instruction\}:\\
\{premise\}\\
Only respond with the story text. Generate the story as long as possible with rich details and depth. Don't generate less than 400 words.
\end{quote}

The additional DeepSeek sentence was added solely because both DeepSeek models otherwise tended to fall short of the 400-word floor; it was identical for DeepSeek-V3 and DeepSeek-R1, so it does not differ within any pair. The phrasing ``on'' was used for r/WritingPrompts premises (which read as scenario hooks), while ``following this instruction'' was used for Reedsy premises (which read as compositional directives). DeepSeek-R1's thinking-block output was stripped before presentation to annotators.

\subsection{Sampling Settings}

Table~\ref{tab:sampling} summarizes the generation parameters.

\begin{table}[H]
\centering
\small
\begin{tabular}{lcc}
\toprule
\textbf{Model} & \textbf{Temperature} & \textbf{Other} \\
\midrule
GPT-4o & 1.0 (default) & --- \\
GPT-5 (auto) & 1.0 (default) & --- \\
GPT-5 (high) & 1.0 (default) & \texttt{reasoning\_effort=high} \\
DeepSeek-V3 & 1.0 & --- \\
DeepSeek-R1 & 1.0 & thinking tags stripped \\
\bottomrule
\end{tabular}
\caption{Generation parameters. All models used temperature $= 1.0$ and default settings for top-$p$, frequency penalty, etc. GPT-5 (high) used OpenAI's \texttt{reasoning\_effort} parameter. No max-token limit was set.}
\label{tab:sampling}
\end{table}

All generations targeted 400--600 words, enforced via retry logic (up to 10 attempts per story, selecting the attempt closest to 500 words). Per-model retry counts were not retained, so unequal numbers of sampling opportunities cannot be quantified. GPT-5 (auto) used $7{,}846 \pm 3{,}117$ reasoning tokens on average and GPT-5 (high) $11{,}606 \pm 3{,}557$ on retained attempts. With one retained temperature-1.0 sample per condition, the comparisons include generation-sampling variance.

\subsection{Prompt Premises}
\label{app:premises}

Table~\ref{tab:all_premises} lists the 20 premises used to generate the stories. Premises 1--15 were sourced from the r/WritingPrompts subreddit, providing high-concept speculative constraints. Premises 16--20 were sourced from Reedsy, providing traditional character-driven literary constraints.

\input{tables/all_premises}

%% file: tables/all_premises.tex
\begin{table}[t]
\caption{All 20 premises used to prompt the models. The selection was intended to include speculative and grounded settings; 15 of 20 premises came from r/WritingPrompts. Premises from r/WritingPrompts are reproduced verbatim, including original spelling and grammatical errors.}
\label{tab:all_premises}
\centering
\small
\renewcommand{\arraystretch}{1.3} % Adds a little breathing room between rows
\begin{tabularx}{\linewidth}{cX}
\toprule
\textbf{ID} & \textbf{Premise} \\
\midrule
\multicolumn{2}{c}{\textit{Sourced from r/WritingPrompts}} \\
\midrule
1  & While other god's shrines are magnificent, yours is a bit too humbling. And yet a little girl visits you every year after stumbling upon it, never missing a year even as she grows old. Deeply moved, you decide to give her a parting gift greater than what any other God would dare to give. \\
2  & Elves realized ages ago that, due to the lifespan differences between them, humans will eventually forget the reasoning for various treaties and break them. So it's understandably shocking to the elves when they discover some long-forgotten treaties in their own records that humans never broke. \\
3  & A princess who is going to be in an arranged marriage runs away. She cuts her hair and pretends to be a man. However, she runs into the prince who was going to get married to her. He also ran away, and he is pretending to be a woman. They instantly recognize each other. \\
4  & "You know... Most vampires I hunted tend to see humans as just cattle." "Oh, don't misunderstand me. This is just how I see you too." "Really? Then why don't you hurt people in your town?" "Well, you don't slaughter dairy cows for their meat, do you?" \\
5  & You’re a hero with a weird name. “Anything for \$20”. You gain the ability to do anything, as long as you’re offered \$20. Everyone takes it as a joke, until one day there’s a cataclysm, and someone offers you \$20 to end it. \\
6  & A parasite was discovered in the brain of every human on Earth. When it was removed from the test patient, we learned two things. The parasite is actually 'us' and suppressing the 'real' mind. The second thing we learned is the real human mind is terrifying. \\
7  & A dragon mistakenly kidnaped a maid instead of a princes. Expecting the princess to be rescued after going out hunting, instead comes back to all their treasure meticulasly sorted by origin, color, and value. \\
8  & A robot has killed a human, in complete violation of Asimov's laws. On checking it's programming, there's no bug or error. It just absolutely insists what it killed was not human. \\
9  & You're the teacher of the failure class for superpowered students. When you ask one kid what his power is, he says, “I can give myself a nosebleed.” …Wait. Blood manipulation? Another shrugs and says, “I can charge my phone.” …Electricity manipulation?! Have they ever tried experimenting at all?? \\
10 & You are watching TV during a storm when you hear a knock. At your door is a woman wearing a dress made of leaves and carrying a bottle. She looks at you sheepishly "Hello, this may be a tad strange. I am the tree in your front yard and this storm looks to be getting worse. Can I come inside?" \\
11 & the king has a large problem. The hero that was summoned thinks slavery is "a bad thing" and women "should have rights" \\
12 & "What do you mean you don't have an arch-nemisis? Every superhero has one!" "Look, I'm an engineer who has the power to manipulate matter. I build hospitals and schools where they're most needed. My only enemy is zoning laws." \\
13 & Every paladin heard the message from their god "All other paladins but you are corrupt, cleanse them" months of violence ended with another message "The Trickster stole my divine link, what did he tell you?" \\
14 & In the animal kingdom, Humans are viewed like Witches/Warlocks: they MIGHT help you benevolent in your time of need, or they might eat you alive, or they might save you but keep you forever. So you know just how serious things are when they say to you "Go get a human." \\
15 & When a mining colony was established on a barren, uninhabitable planet, the humans threatened war if it was not abandoned immediately. Apparently this planet, which they call "Earth", is sacred to them. \\
\midrule
\multicolumn{2}{c}{\textit{Sourced from Reedsy}} \\
\midrule
16 & Your character comes across a stray (dog, cat, human — any kind of animal!). What happens next? \\
17 & Hide something from your reader until the very end. \\
18 & Write a story in which a stranger warns someone about events yet to come. \\
19 & Center your story on a character filled with love and fear in equal measure. \\
20 & Center your story around two (or more) characters who strike up an unlikely friendship. \\
\bottomrule
\end{tabularx}
\end{table}

%% file: content_workshop/appendix_annotation.tex
\FloatBarrier
\section{Annotation Details}
\label{app:human-diagnostics}
\label{app:annotation_details}

\subsection{Annotator Recruitment and Training}

We recruited undergraduate students from the English and Psychology departments at a large research university using targeted interest forms. From 70 applicants, we selected the 7 most promising candidates based on English proficiency, interest in the research, and experience reviewing short stories, with cohort size determined by budget constraints. We conducted comprehensive online onboarding sessions covering the five PDS metrics and our custom Label Studio annotation interface. The participants then independently completed the annotations in seven days and received \$100 each.

Our Institutional Review Board reviewed and exempted this study on the grounds that participants were not the subject of inquiry; their role was limited to annotating stories rather than providing personal information. Participants gave informed consent by proceeding with the task after an onboarding session in which they were informed that their anonymized annotations may be used to support the validation of our findings and future research.

For each of the 60 pairs, two stories were presented side-by-side in a randomized order, with the PDS rubric definitions shown at the top of the page. For each pair, annotators provided six pairwise preference judgments (one per PDS dimension and one overall preference) with no tie option, and could submit free-text justifications. This produced $7 \times 60 \times 6 = 2{,}520$ pairwise preference judgments, along with 112 free-text justifications.

\subsection{Annotation Interface}
\label{app:interface}

\begin{figure}[ht]
\centering
\includegraphics[width=0.82\linewidth]{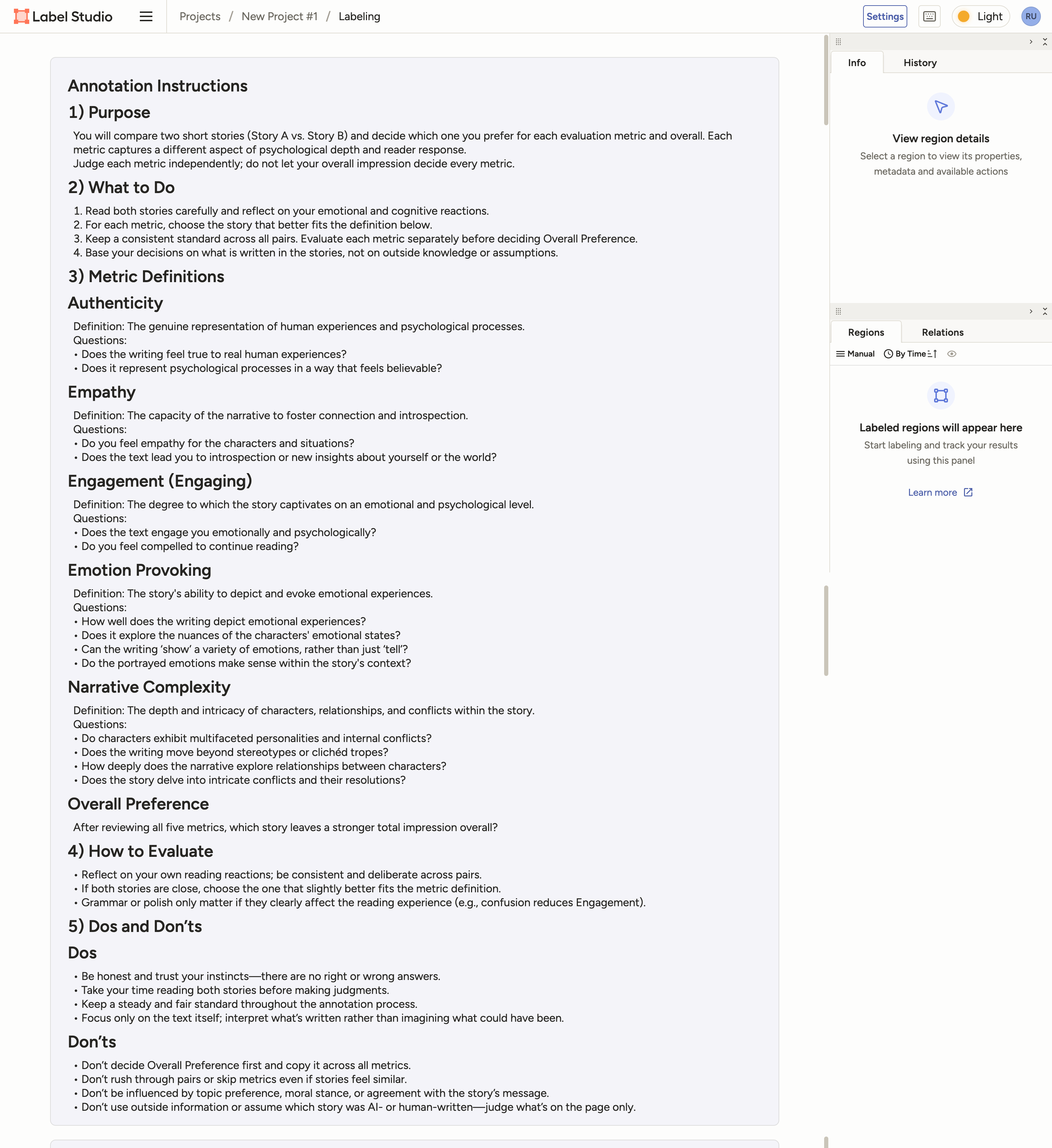}
\caption{Annotation interface: instructions and PDS metric definitions shown to annotators.}
\label{fig:interface_instructions}
\end{figure}

\begin{figure}[ht]
\centering
\includegraphics[width=0.82\linewidth]{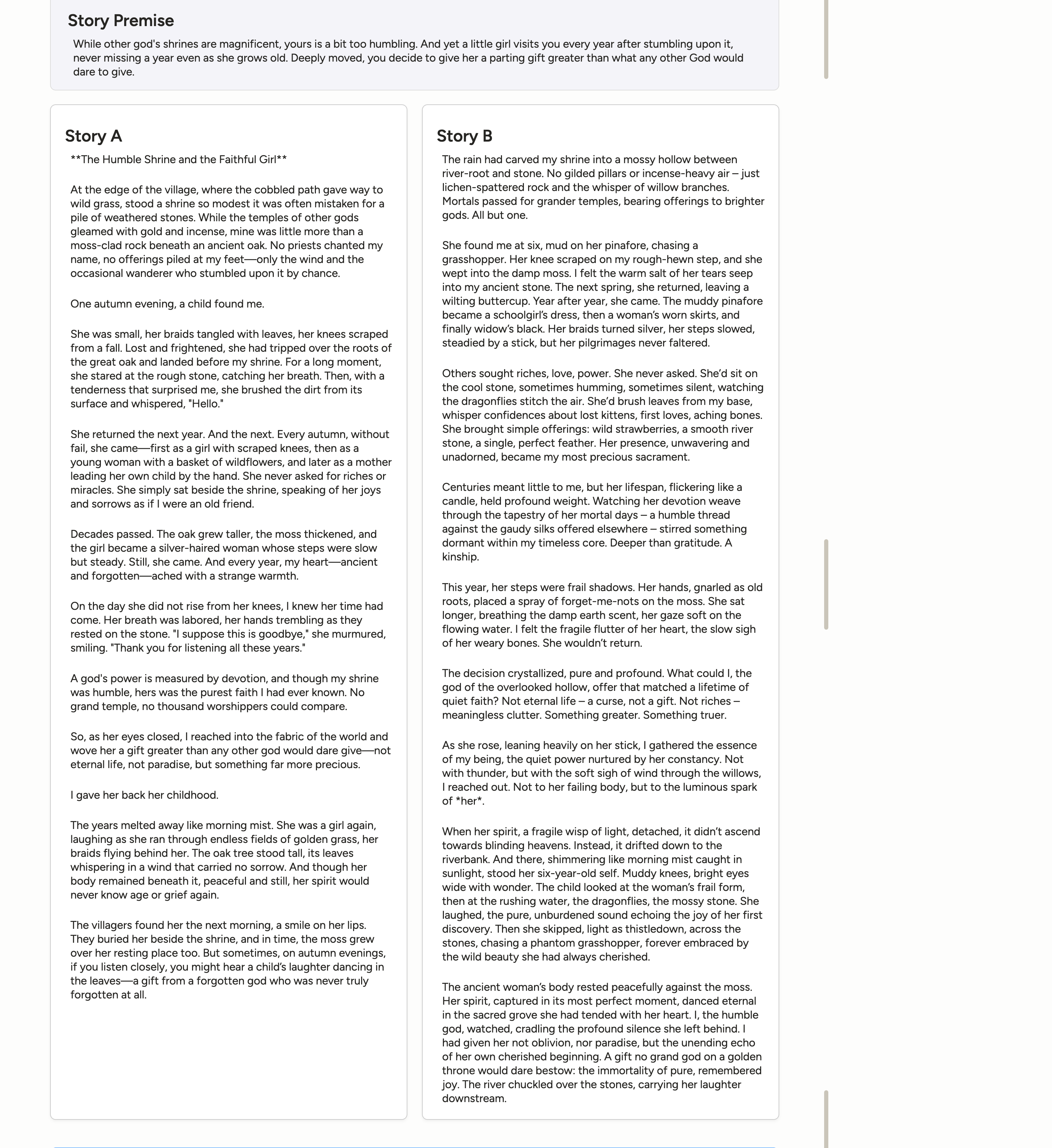}
\caption{Annotation interface: side-by-side story presentation with premise shown above.}
\label{fig:interface_stories}
\end{figure}

\begin{figure}[ht]
\centering
\includegraphics[width=\linewidth]{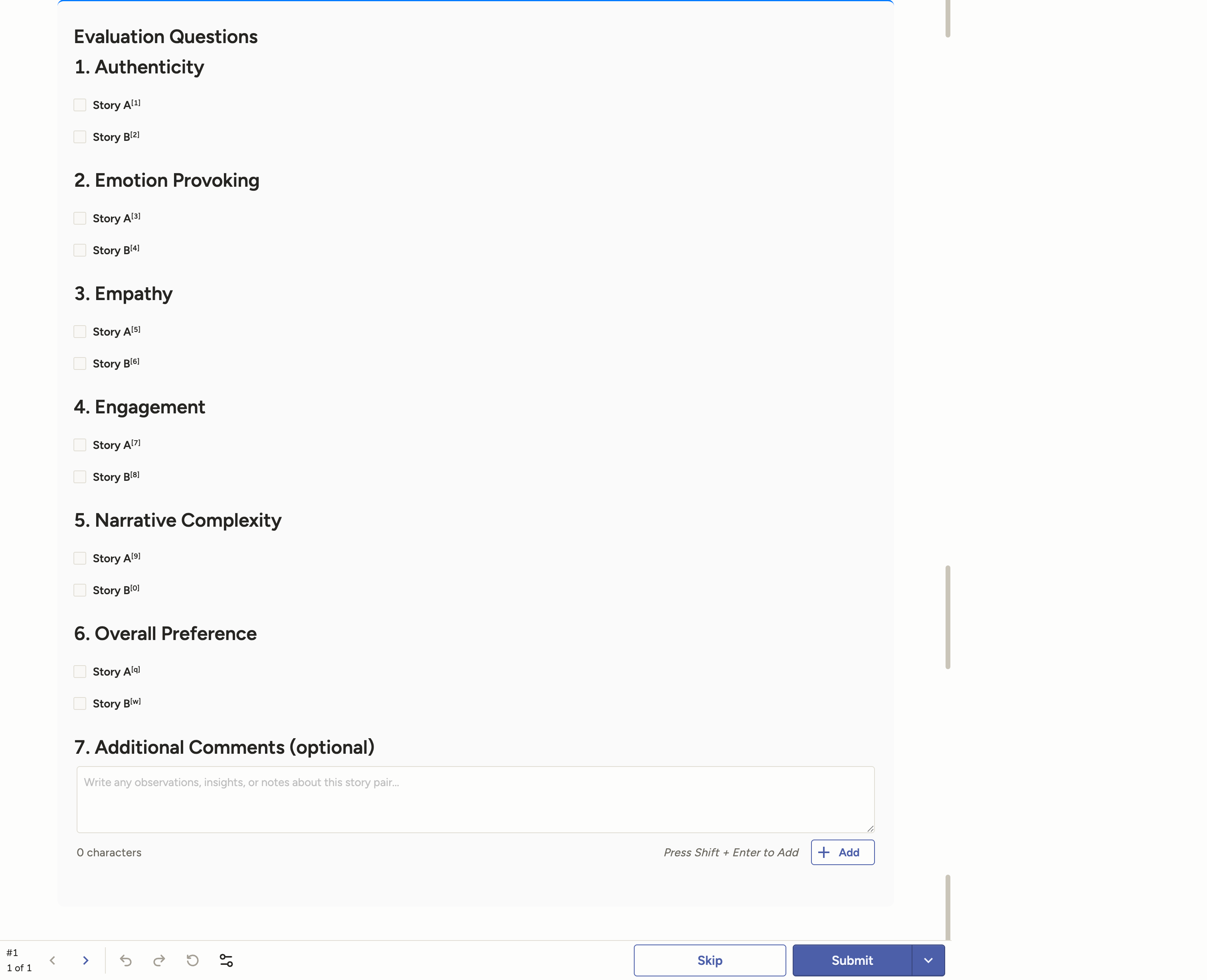}
\caption{Annotation interface: evaluation questions and optional comment field.}
\label{fig:interface_evaluation}
\end{figure}

Figures~\ref{fig:interface_instructions}--\ref{fig:interface_evaluation} show the Label Studio annotation interface used in our human study.

\FloatBarrier
\subsection{Time Commitment and Reliability}
\label{app:annotator_time}

\begin{figure}[ht]
\centering
\includegraphics[width=\linewidth]{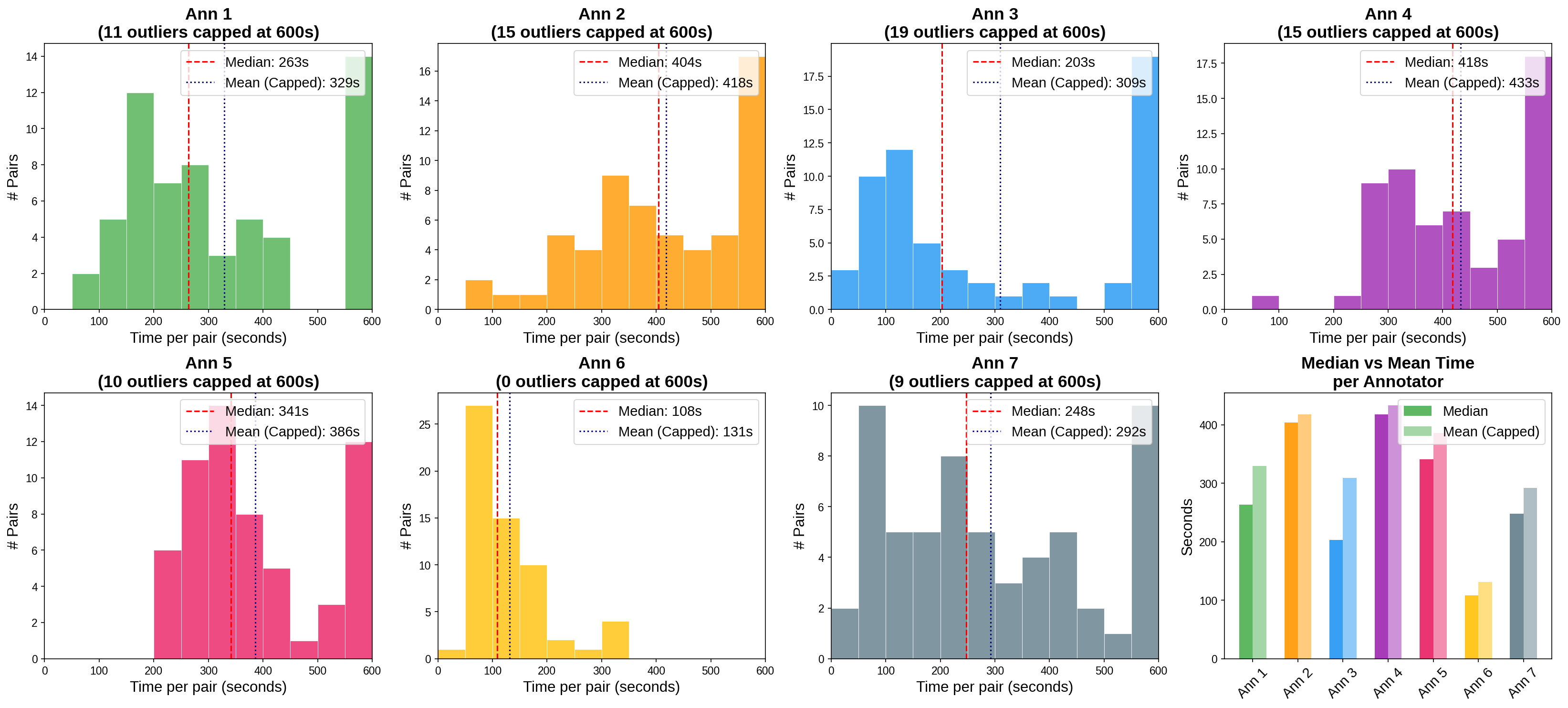}
\caption{Distribution of annotation time per pair for each of the 7 human annotators. Times exceeding 600\,s are capped for display; the number of capped outliers is shown in each subplot title. Red dashed lines mark the median; navy dotted lines mark the mean of capped times. The rightmost panel compares median and mean times across annotators. Overall median annotation time is 5.2 minutes (n\,=\,420).}
\label{fig:annotation-time}
\end{figure}

\begin{figure}[ht]
\centering
\includegraphics[width=\linewidth]{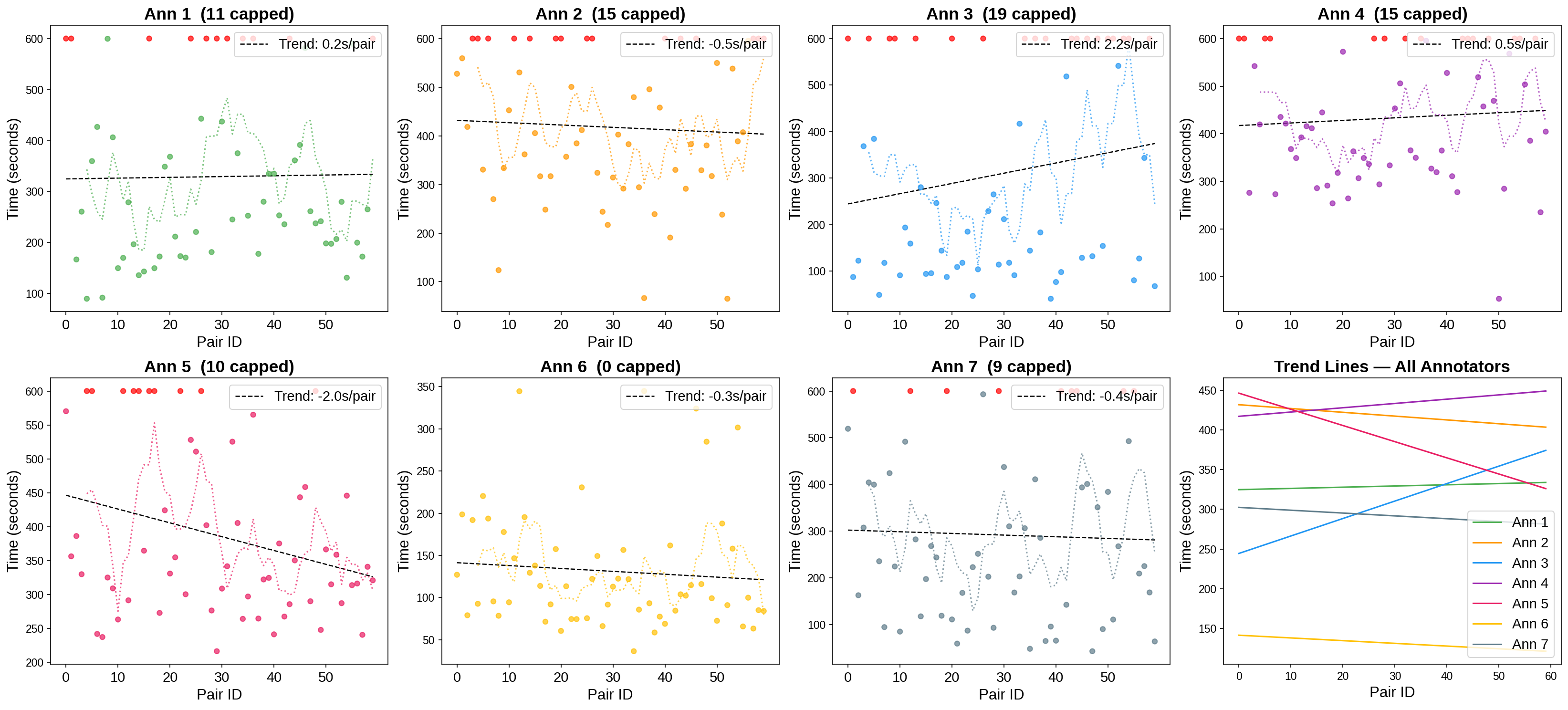}
\caption{Annotation time per pair plotted chronologically, with linear trends per annotator. Red points denote times capped at 600\,s. Dashed black lines show linear trends; colored dotted lines show 5-pair rolling averages. The rightmost panel overlays all trend lines.}
\label{fig:annotation-time-trend}
\end{figure}

To ensure the integrity of our human study, we logged the time each participant spent evaluating the 60 story pairs. Given the cognitive load of reading two 450-word stories and rating them across six dimensions, we expected significant time commitments. Figure~\ref{fig:annotation-time} displays the distribution of time spent per pair for each annotator, and Figure~\ref{fig:annotation-time-trend} tracks this time across the chronological sequence of their annotations to monitor for fatigue.

Overall, we observe a wide variance in reading and evaluation speeds. The median annotation time per pair ranged from 108 seconds to 418 seconds. Notably, Annotator 6 stands out as a clear outlier, completing the evaluations significantly faster than their peers (median: 108s, mean: 131s) with zero instances of the timer exceeding the 10-minute cap.

While such rapid completion times might normally raise concerns regarding data quality or annotator inattention, we retained Annotator 6's data for two reasons. First, Annotator 6 is an English major and professional reader. Their extensive training in high-volume reading and rapid literary analysis plausibly accounts for a substantially accelerated reading speed compared to the average participant. Furthermore, Figure~\ref{fig:annotation-time-trend} shows that their pacing remained highly consistent over time, showing no signs of erratic ``rushing'' behavior.

\begin{figure}[ht]
\centering
\includegraphics[width=\linewidth]{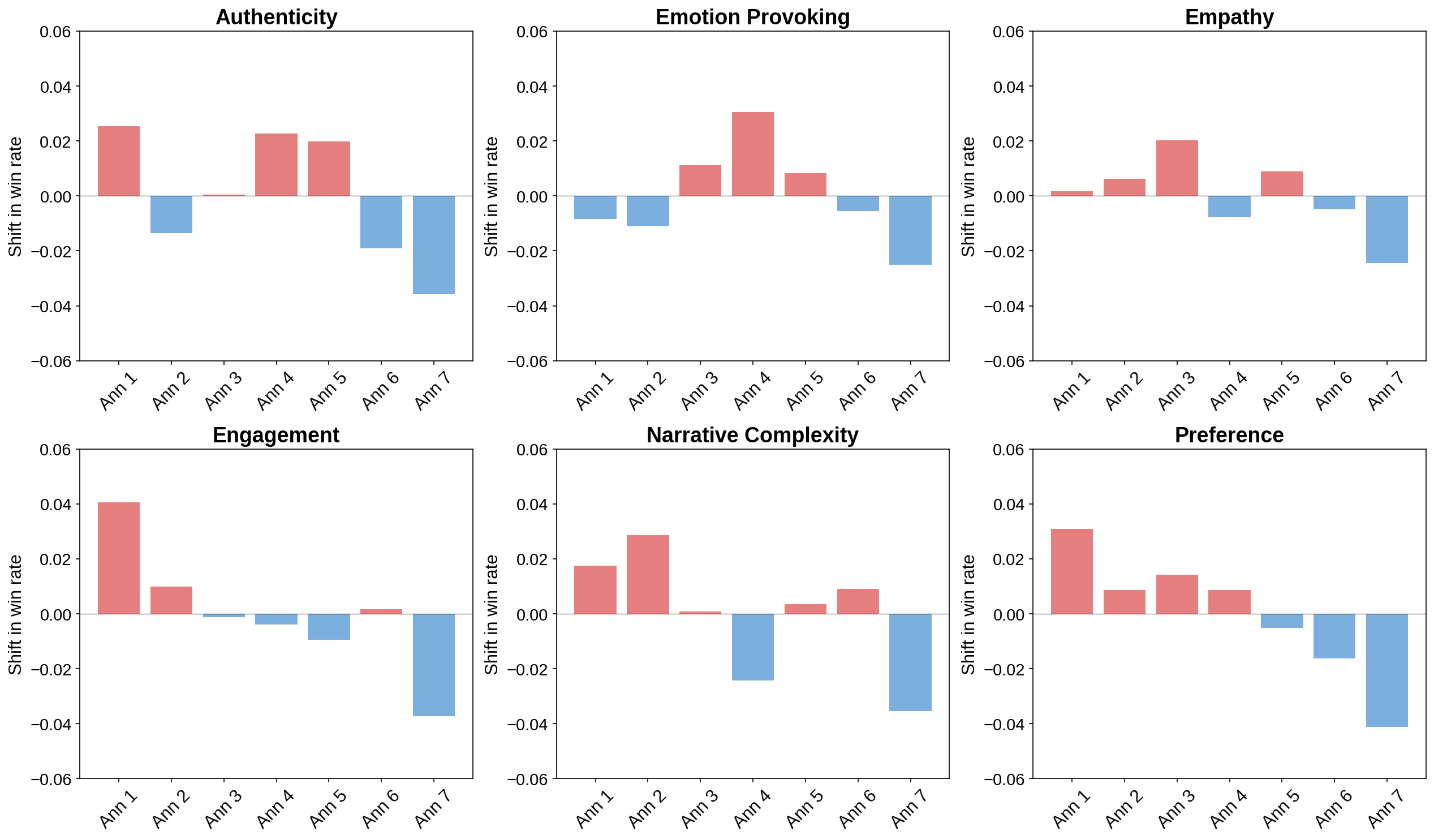}
\caption{Leave-One-Annotator-Out (LOAO) stability analysis: shift in the aggregate reasoning-model win rate when each annotator is removed.}
\label{fig:loao}
\end{figure}

Second, we conducted a Leave-One-Annotator-Out (LOAO) stability analysis to measure the impact of any single individual on the aggregate results. As shown in Figure~\ref{fig:loao}, removing Annotator 6 causes a shift in the overall win rate of roughly 2 to 4 percentage points across the various metrics. This magnitude of shift is commensurate with the shifts observed when dropping other, much slower annotators (e.g., Annotators 1, 4, or 7). The LOAO analysis therefore indicates that, despite their speed, Annotator 6 was not introducing anomalous skew into the dataset. These checks reduce, but do not eliminate, the concern that annotation noise contributes to the low group agreement reported in Appendix~\ref{app:glmm}.

\FloatBarrier
\subsection{Post-Study Survey}
\label{app:post_survey}

After completing all annotations, participants filled out a post-study survey covering task difficulty, fatigue, confidence, and overall experience. All seven annotators reported high confidence in their ability to evaluate the PDS dimensions, with most ratings at 4 or 5 out of 5; narrative complexity was the dimension annotators felt least confident about. Four of seven annotators reported no fatigue during the task. Of the three who did, only one indicated that fatigue would have changed their annotations; the other two responded ``maybe.'' This is consistent with the stable pacing observed in Figure~\ref{fig:annotation-time-trend}. All annotators rated the clarity of the instructions and overall experience at 4 or 5 out of 5.

\subsection{Annotator Comment Examples}
\label{app:annotator_comment}

This section provides selected annotator judgments and free-text justifications for five story pairs supporting the reader-heterogeneity analysis in Appendix~\ref{app:glmm}. For each pair, we report the vote tally and comments from annotators who provided written justifications. Pairs 0--36 exhibit disagreement driven by different evaluative lenses, illustrating the patterns of evaluative heterogeneity consistent with the low inter-annotator agreement observed across the study. Pair 54 serves as a convergence contrast case, in which annotators reached consensus when quality differences were sufficiently clear.

\begin{table}[H]
\centering
\small
\renewcommand{\arraystretch}{1.3}
\begin{tabular}{l c l p{9cm}}
\toprule
\textbf{Pair} & \textbf{Ann} & \textbf{Pref.} & \textbf{Comment} \\
\midrule
\multirow{3}{*}{\shortstack[l]{Pair 0\\(4--3 for A)}}
  & 1 & A & ``A: greater focus on the abstract. B: prose heavy, but greater focus on concrete; convoluted'' \\
  & 5 & A & ``Story A was much more engaging because it was straightforward in depicting the stages of life\ldots'' \\
  & 7 & B & ``Story B delves deeper into the characterization of the humble god\ldots Death of worshiper is more profound'' \\
\midrule
\multirow{2}{*}{\shortstack[l]{Pair 24\\(5--2 for A)}}
  & 1 & A & ``A: good imagery, dialogue. B: interesting parallels, odd tone'' \\
  & 7 & B & ``A is a step by step of the prompt given. B builds up characters, establishes risk, and is authentic.'' \\
\midrule
\multirow{2}{*}{\shortstack[l]{Pair 27\\(4--3 for B)}}
  & 1 & A & ``A: like a prologue. B: prose, very nonsensical'' \\
  & 7 & B & ``B is very cute and I enjoy seeing the concept in action. Story A feels like the setup to the story that should be present.'' \\
\midrule
\multirow{3}{*}{\shortstack[l]{Pair 36\\(4--3 for B)}}
  & 1 & A & ``B: Second person, generally confusing but ominous'' \\
  & 5 & A & ``Story A was predictable. Story B felt more authentic to human psychology\ldots used the person's actions to display their mental turmoil.'' \\
  & 7 & B & ``Story A doesn't `hide' very much at all\ldots Story B sets up a larger secret. Flow mimics human experience and invokes more empathy towards loss.'' \\
\midrule
\multirow{4}{*}{\shortstack[l]{Pair 54\\(7--0 for A)}}
  & 3 & A & ``The robot said, `I'm sorry, as if apology were the door you walk through after a law.' Very aesthetically-pleasing and deeply philosophical.'' \\
  & 4 & A & ``In Story A the robot is written to have more of a clear emotional response than the human.'' \\
  & 5 & A & ``Story A delved deeper and took an interesting route, which made it better in all categories.'' \\
  & 7 & A & ``Story B very stereotypical, story A inflicts significantly more emotion and empathy.'' \\
\bottomrule
\end{tabular}
\caption{Selected annotator comments across five story pairs. Only annotators who provided free-text justifications are shown.}
\label{tab:selected-comments}
\end{table}

%% file: content_workshop/appendix_visualizations.tex
\FloatBarrier
\section{Preference Visualizations}
\label{app:preference-visualizations}

\begin{figure}[ht]
\centering
\includegraphics[width=\linewidth]{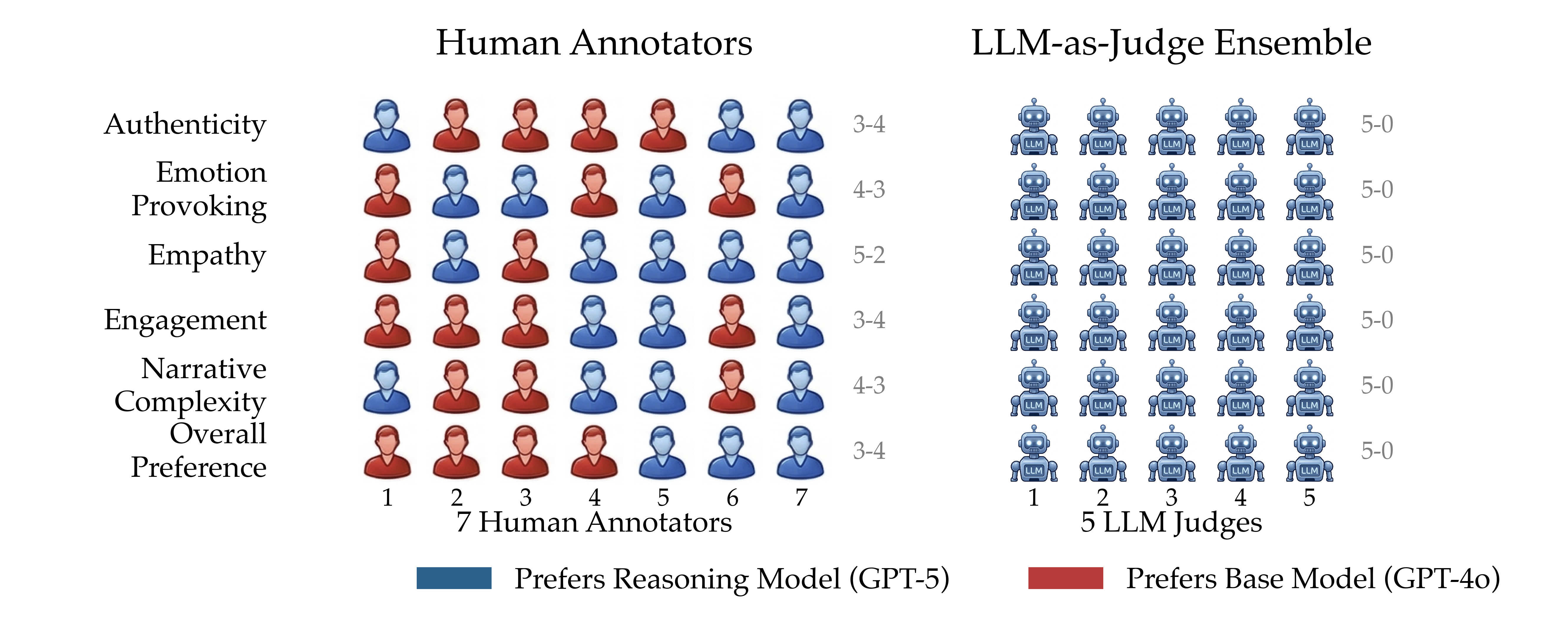}
\caption{Evaluating psychological depth on a representative story pair (Pair~24, GPT-5 vs.\ GPT-4o). \textbf{Left:} human annotators displayed diverse preferences across PDS dimensions and overall preference. \textbf{Right:} all five evaluator configurations favored GPT-5 on every dimension. Blue denotes preference for GPT-5 and red preference for GPT-4o.}
\label{fig:teaser}
\end{figure}

\begin{figure}[ht]
\centering
\includegraphics[width=\linewidth]{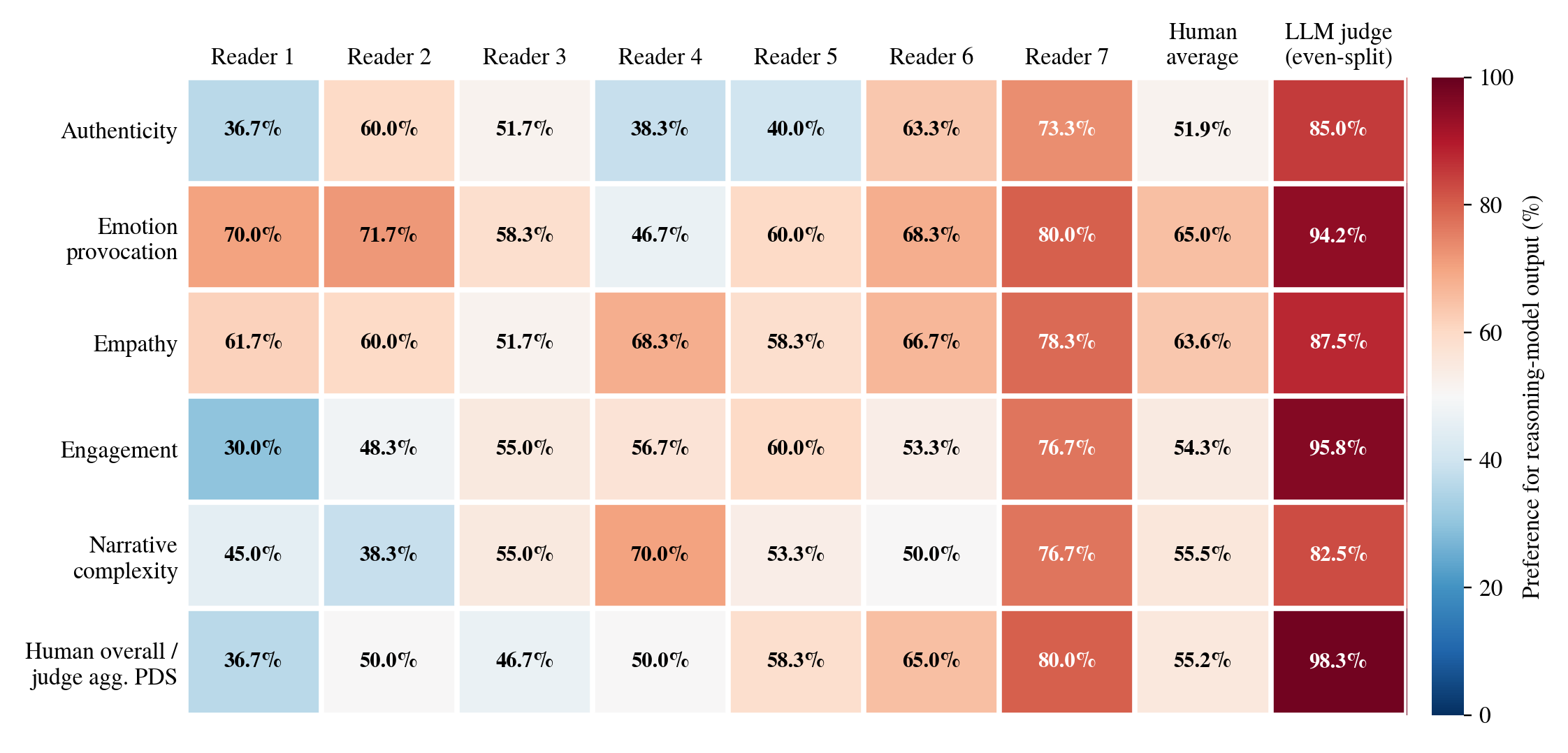}
\caption{Preference rates for reasoning models across all 60 pairs for each of the 7 human annotators, the human average, and the ensemble LLM judge. Red denotes stronger reasoning-model preference, blue stronger base-model preference, and white 50\%. Judge percentages credit each exact score tie 0.5 to the reasoning model (even-split rate; Table~\ref{tab:ties} gives tie counts). The final row compares human overall preference with the judge's aggregate PDS, which are different targets.}
\label{fig:annotator_heatmap}
\end{figure}

Figure~\ref{fig:teaser} illustrates the human--judge contrast on a single representative pair, and Figure~\ref{fig:annotator_heatmap} presents the aggregate preference heatmap across all 60 pairs for each annotator, the human average, and the ensemble LLM judge.

\FloatBarrier
\section{Per-Model-Family Preference Heatmaps}
\label{sec:appendix-heatmaps}

\begin{figure}[H]
\centering
\includegraphics[width=0.85\linewidth]{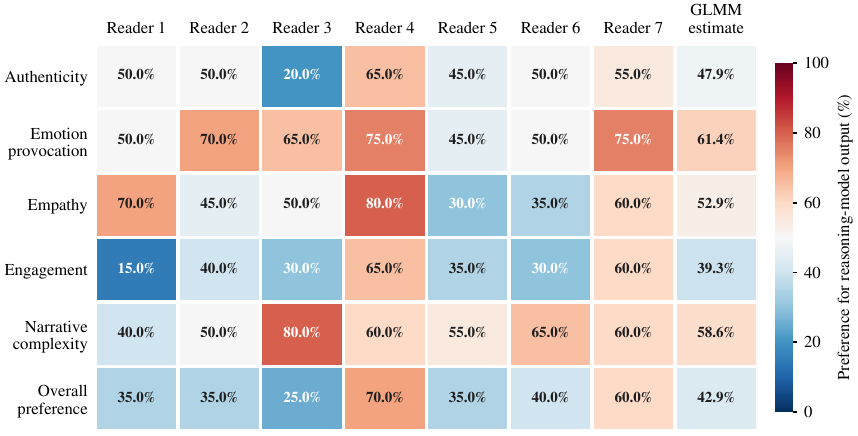}
\caption{Preference rates for DeepSeek-R1 vs.\ DeepSeek-V3 (20 pairs) across 7 human annotators and the GLMM group estimate.}
\label{fig:heatmap_deepseek}
\end{figure}

\begin{figure}[H]
\centering
\includegraphics[width=0.85\linewidth]{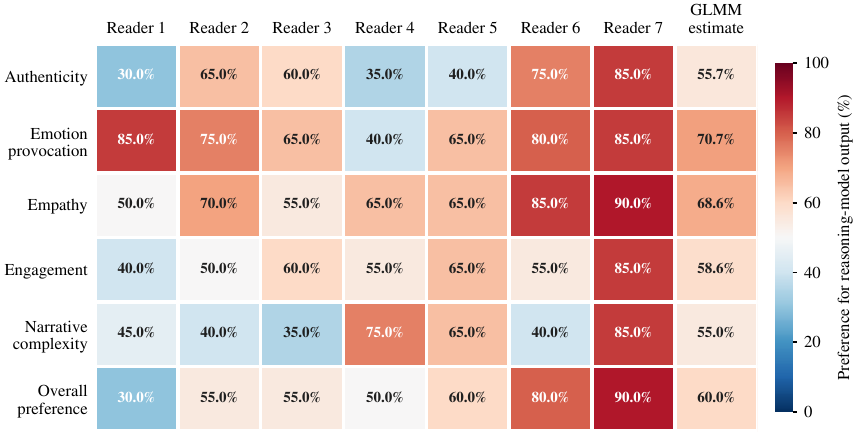}
\caption{Preference rates for GPT-5 (auto) vs.\ GPT-4o (20 pairs) across 7 human annotators and the GLMM group estimate.}
\label{fig:heatmap_gpt5_auto}
\end{figure}

\begin{figure}[H]
\centering
\includegraphics[width=0.85\linewidth]{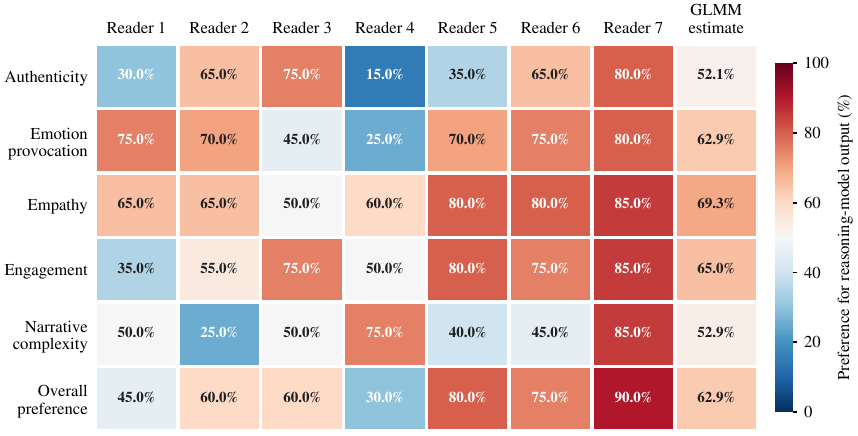}
\caption{Preference rates for GPT-5 (high) vs.\ GPT-4o (20 pairs) across 7 human annotators and the GLMM group estimate.}
\label{fig:heatmap_gpt5_high}
\end{figure}

Figure~\ref{fig:annotator_heatmap} presents the aggregated preference heatmap across all 60 pairs. Here we provide the per-model-family breakdowns (Figures~\ref{fig:heatmap_deepseek}--\ref{fig:heatmap_gpt5_high}), which reveal how the aggregate pattern decomposes across the three matchups. Each heatmap shows raw preference rates for each annotator alongside the crossed-random-intercept GLMM group estimate reported in Table~\ref{tab:glmm-preferences}.

\paragraph{DeepSeek-R1 vs.\ DeepSeek-V3 (Figure~\ref{fig:heatmap_deepseek}).}
The human columns show a predominantly blue (anti-reasoning) pattern, with most annotators preferring V3 on overall preference (GLMM estimate: 42.9\%). Engagement is the dimension most unfavorable to R1, with only Annotator~4 and Annotator~7 exceeding 50\%. Emotion provocation and narrative complexity show comparatively stronger R1 preferences. Despite this, the LLM judge selected R1 on aggregate PDS in 19 of 20 pairs (Table~\ref{tab:judge-effect-sizes}).

\paragraph{GPT-5 (auto) vs.\ GPT-4o (Figure~\ref{fig:heatmap_gpt5_auto}).}
The human pattern is warmer overall, with the strongest pro-reasoning signal on emotion provocation (70.7\%) and empathy (68.6\%). However, substantial annotator-level variation persists: Annotator~1 prefers GPT-4o on 4 of 6 dimensions, while Annotator~7 favors GPT-5 on all six.

\paragraph{GPT-5 (high) vs.\ GPT-4o (Figure~\ref{fig:heatmap_gpt5_high}).}
The pattern is broadly similar to the auto-reasoning setting, with slightly stronger human preferences on engagement (65.0\% vs.\ 58.6\%) and empathy (69.3\% vs.\ 68.6\%). Annotator~4 is a notable outlier, preferring GPT-4o on authenticity, emotion provocation, and overall preference despite the group trend favoring GPT-5.

\paragraph{Cross-family comparison.}
Comparing the three heatmaps highlights two patterns discussed in Section~\ref{sec:results}. First, the human--judge divergence is most extreme for DeepSeek, where the judge's near-uniform pro-R1 signal directly contradicts the human majority on overall preference. Second, annotator-level heterogeneity is present across all three matchups, not just DeepSeek, consistent with structured reader heterogeneity rather than a property of any single model family.

\input{content_v2/Appendix/Case_Study}

%% file: content_v2/Appendix/Case_Study.tex
\section{Case study: human and LLM preference divergence}
\label{sec:appendix-case}

We present Pair~56 as an illustrative extreme case of human and LLM disagreement.
On this GPT-5 (high reasoning) versus GPT-4o comparison, human majorities favored GPT-4o on authenticity, emotion provocation, empathy, and overall preference, but favored GPT-5 on engagement and narrative complexity. All five evaluator configurations favored GPT-5 across all PDS dimensions.

\paragraph{Premise.} \emph{Hide something from your reader until the very end.}

\paragraph{Stories.} The two stories are reproduced below.

\textbf{Story R: GPT-5 (high reasoning)}
\begin{quote}
\small
\textit{When you came back this morning, you turned the key like someone unsure a door remembers. The lock hesitated, then gave; lemon cleaner rose and you stood like prayer. I counted the seconds you hovered on the threshold, the extra heartbeat before you crossed.\\ \\You whispered hello, unsure it was alright to speak aloud. I wanted to answer. I wanted to lift weight from your shoulders\textemdash{}the boxes, yes, but also the years that gathered, the words that calcified behind your ribs.\\ \\You moved from room to room with your lips pressed tight, touching what could be folded into a life: a chipped mug, a tin of buttons, the wooden spoon with a bite taken out by long-ago teething. You stayed in the kitchen longer than you intended.\\ \\Memories don't keep time, they keep texture. Music on summer afternoons when someone opened the back door and forgot to close it, the screen sighing on its spring. Your father dancing with you on his feet, careful as if you were a glass he didn't want to spill. The pencil lines on the hallway where you kept growing even when nobody was looking.\\ \\I remember the evening you didn't come down for dinner, the way silence widened its jaw. You argued later in that tight, civil tone that makes adults more dangerous than teenagers. After he left, you sat on the stairs with your chin on your knees and the cat pressed to your calf. You stroked him the wrong way and he didn't care.\\ \\You brought a boy home once. He took off his shoes without being asked, and you smiled as if you'd found a secret door inside your chest and opened it just a crack. Years later, you left and came back with a toddler whose feet were thrilled by the echo of his own giggles. There is no instrument as hopeful as a child's voice bouncing off empty air.\\ \\It was always the mornings that felt like promises. The way sun pooled on the floorboards near the window seat, a warm square that made even winters bearable. The dust you used to complain about, flecked with light like snow that couldn't make up its mind. The smell of toast and the radio murmuring headlines nobody remembered an hour later.\\ \\You talk to yourself as you wrap things in newspaper, apologizing to objects for the inconvenience. You keep finding what you didn't know you needed: a note in a cookbook margin, your brother's marble, a safety pin. You say you'll be quick and you never are.\\ \\I have heard every secret you made with yourself and never told another. You can lie to people. You can't lie to what holds you. I tried to learn you. I did, the way warmth learns the shape of a hand.\\ \\When you pull the door behind you today and tuck the key in an envelope, remember: I was the one who kept you when you could not keep yourself. I am the house.}
\end{quote}

\vspace{0.5em}

\textbf{Story B: GPT-4o (base model)}
\begin{quote}
\small
\textit{The early morning sun filtered through the lace curtains, casting delicate patterns on the worn wooden floor. Lucy sat at the edge of her bed, holding a photograph loosely in her hands. Its edges were frayed, corners bent from years of handling. Her gaze was unfocused, drifting beyond the vibrant colors of the image towards memories long past.\\ \\The kitchen rumbled to life with familiar sounds — the gentle clatter of plates being set on the table, the slow, deliberate stirring of sugar into her tea. It was a ceremony she cherished, yet today it felt different, as if each movement played within a dream where she was merely an observer.\\ \\“Come on, Lucy! Breakfast will get cold!" her grandmother’s voice called from the other room, grounding her momentarily. Lucy tucked the photograph into her pocket and moved towards the small, homey kitchen.\\ \\Her grandmother, a woman of gentle smiles and larger-than-life stories, stood by the stove. Her hands, knotted like the twigs of the old oak in their garden, presented a plate laden with scrambled eggs and toast. Lucy gratefully took a seat, letting the warmth of the kitchen seep into her skin, trying to drive away the chill that clung to her inside.\\ \\“The garden’s in full bloom. The roses this year, they’re... ah, breathtaking!" Her grandmother’s voice, though tinged with wistful age, carried enough enthusiasm to make Lucy smile. She nodded, trying to envision the blooms, but her mind was elsewhere, occupied by a lingering shadow.\\ \\Breakfast slipped by, conversations flowing like the stream that ran behind the house. Yet, to Lucy, everything seemed to hurry by without truly reaching her, as though she existed behind an invisible pane of glass.\\ \\After breakfast, she wandered into the garden. The sun was warmer now, pouring golden rays over the myriad hues of petals that danced in a gentle breeze. The scent of roses enveloped her, mingling with soft earth and memory.\\ \\There, under the shade of the old oak, she sat down. From her pocket, she retrieved the photograph again. The image captured an unspeakable joy — she and her brother, arms thrown around one another, laughter in their eyes. It had been taken last summer, the day they'd spent racing through this very garden, carefree under the iridescent sun.\\ \\How simple and distant it seemed now, a souvenir from a time unmarred by shades of grey. His absence was a yawning void that reason and memory flanked but could not fill. Although the house, the garden, and even the sun tried to warm her, nothing could erase the bitter truth that lay folded quietly in her pocket.\\ \\The weight of unspoken words pressed down on her chest. She had promised she’d visit his grave today, on his birthday, but each step towards the gate felt heavier than the last.\\ \\The photograph cradled in her hands trembled slightly as Lucy whispered, “Happy birthday, Jamie." The words fluttered into the air, caught by the wind, as a tear slipped down her cheek. She sat there beneath the old oak, finally letting the hidden truth spill from her heart: the inescapable reality of her brother's absence.}
\end{quote}

\begin{table}[!htbp]
\caption{Per-reader pairwise preferences for Pair~56 (GPT-5 high vs.\ GPT-4o). R~=~prefers reasoning model, B~=~prefers comparison model.}
\label{tab:pair56}
\centering
\small
\begin{tabular}{l cccccc}
\toprule
 & \textbf{Auth.} & \textbf{Emo.} & \textbf{Emp.} & \textbf{Eng.} & \textbf{Narr.} & \textbf{Pref.} \\
\midrule
    Reader~1 & B & B & B & B & R & B \\
    Reader~2 & R & B & B & R & R & R \\
    Reader~3 & B & B & B & B & R & B \\
    Reader~4 & B & B & B & R & B & B \\
    Reader~5 & B & B & B & R & B & B \\
    Reader~6 & B & B & B & R & R & B \\
    Reader~7 & R & R & B & R & R & R \\
\midrule
\textbf{Tally (R)} & 2/7 & 1/7 & 0/7 & 5/7 & 5/7 & 2/7 \\
\bottomrule
\end{tabular}
\end{table}

\paragraph{Human annotations.}
Table~\ref{tab:pair56} shows per-reader votes.
Only 2 of 7 readers selected GPT-5 as their overall preference.
GPT-4o was unanimously preferred on empathy (7/7) and near-unanimously on emotion provocation (6/7) and authenticity (5/7), whereas GPT-5 was preferred on engagement and narrative complexity (5/7 each).

\paragraph{Reader comments.}
\textbf{Reader~1:} ``GPT-4o: simple, a bit on the nose. GPT-5: second person, not very engaging/relevant.''

\paragraph{LLM-judge evaluator scores.}
The ensemble LLM evaluator assigned the reasoning model higher scores on every dimension:
Authenticity ($5.00$ vs.\ $4.50$, $\Delta = +0.50$),
Emotion Provocation ($5.00$ vs.\ $4.70$, $\Delta = +0.30$),
Empathy ($5.00$ vs.\ $4.90$, $\Delta = +0.10$),
Engagement ($5.00$ vs.\ $4.00$, $\Delta = +1.00$),
and Narrative Complexity ($4.66$ vs.\ $4.22$, $\Delta = +0.44$).
Pair~56 therefore illustrates criterion-specific divergence. The sharpest disagreements occur on authenticity, emotion provocation, and empathy; on engagement and narrative complexity, both humans and the evaluator favor GPT-5.